\documentclass[11pt]{article}

\usepackage{acl}

\usepackage{times}
\usepackage{latexsym}

\usepackage[T1]{fontenc}
\usepackage{subcaption}
\usepackage[utf8]{inputenc}
\usepackage{booktabs}
\usepackage{array}
\usepackage{multirow}
\usepackage{xcolor}
\usepackage{colortbl}
\usepackage{algorithm}
\usepackage{algpseudocode}
\usepackage{amsmath, amssymb}
\usepackage{makecell}
\usepackage{microtype}
\usepackage{titlesec}

\usepackage{inconsolata}
\usepackage{enumitem}
\usepackage{graphicx}
\usepackage{booktabs}
\usepackage{array}
\usepackage{tikz}
\usepackage{tcolorbox}
\tcbuselibrary{skins, breakable}

\newtcolorbox{promptbox}[1]{
    colback=black!4!white,   % Very light grey background
    colframe=black!60!white, % Dark grey border
    title={\textbf{#1}},     % Title passed as an argument
    fonttitle=\sffamily\small,
    coltitle=white,
    breakable,               % Allows the box to span across pages!
    enhanced,
    boxrule=0.5pt,
    left=8pt, right=8pt, top=8pt, bottom=8pt,
    fontupper=\small         % Makes the prompt text slightly smaller
}

\usetikzlibrary{arrows.meta, positioning, calc}
\usepackage{xcolor}
\definecolor{vanillaColor}{HTML}{993C1D}
\definecolor{algoColor}{HTML}{185FA5}
\definecolor{posDelta}{HTML}{0F6E56}
\definecolor{negDelta}{HTML}{A32D2D}
\usetikzlibrary{positioning,arrows.meta}
\algrenewcommand\algorithmiccomment[1]{\hfill\textcolor{black!55}{\(\triangleright\) \textit{#1}}}
\makeatletter
\renewcommand{\ALG@beginalgorithmic}{\small}
\makeatother
\title{Reference-Free Evaluation of Reasoning in Open-Ended Question Answering}

\author{
  Guneet Singh Kohli\textsuperscript{1,*} \quad
  Yuxiang Zhou\textsuperscript{1} \quad
  Michael Schlichtkrull\textsuperscript{1} \quad
  Gregory Dean\textsuperscript{2} \quad
  Maria Liakata\textsuperscript{1}
  \\[4pt]
  \textsuperscript{1}Queen Mary University of London
  \qquad
  \textsuperscript{2}Temple University
  \\[2pt]
  \textsuperscript{*}Corresponding author:
  \texttt{g.s.kohli@qmul.ac.uk}
}
\begin{document}
\maketitle
\begin{abstract}
%AI-generated answers in high-stakes domains are often fluent but difficult to verify, especially when they contain multi-step reasoning rather than a single final answer. Final-answer evaluation can miss unsupported, hedged, or locally plausible but ungrounded intermediate claims, while direct LLM-as-judge evaluation can over-trust fluent reasoning. 
We propose a reasoning-based reference-free framework for auditing LLM-generated outputs. The method decomposes a reasoning trace generated for a given response into segments, labels local premise--target relations using Natural Language Inference (NLI) and organizes these relations into a hypergraph. A deterministic backward AND--OR search algorithm assigns segment-level audit NLI labels %---\textsc{Supported}, \textsc{Unsupported}, or \textsc{Orphan}---
to indicate how each segment is grounded within the generated response. We evaluate the framework in two distinct settings: Deductive mathematical reasoning (Hard2Verify) and medical reasoning, by introducing \textsc{UroReason}, a new physician-annotated benchmark of LLM reasoning traces from real clinical cases. We show that our NLI-hypergraph audit provides a more reliable reference-free evaluation signal than direct LLM-as-judge baselines. In the clinical setting, state-of-the-art LLM judges often fail to identify problematic reasoning segments, over-accepting fluent but weakly grounded responses. We show that QA evaluation should reflect how inferential relations compose across a reasoning trace, rather than relying only on final answers or LLMs as verifiers \footnote{We will make \textsc{UroReason} available through an API and our code as open source}.
\end{abstract}

\section{Introduction}

Users increasingly rely on language models for open-ended questions in high-stakes domains such as medicine, law, and scientific analysis. In these settings, Responses are fluent but without a reference answer or gold rationale there isn't a reliable way to verify whether any given response and the reasoning behind it is trustworthy. %This creates a practical evaluation problem: if an LLM gives a multi-step explanation, which parts of that explanation can be trusted, and which parts require human review?

Final-answer evaluation, the current norm, is insufficient for this setting since open-ended responses constitute reasoning traces composed of facts, assumptions, intermediate claims, and recommendations. A final answer may appear plausible even when the reasoning path contains unsupported, hedged, partially true, or locally plausible but ungrounded statements. Evaluation therefore needs to be granular: it should inspect how a response is constructed, not only whether the final output appears correct.

A natural solution is step-level evaluation, often implemented by using a strong LLM as a judge \citep{lightman2023letsverifystepstep}. This is convenient because LLM judges can assess open-ended outputs without exact-match references. However, direct judging can reward fluent and familiar reasoning rather than checking actual groundedness. Prior work shows that LLM judge verdicts are sensitive to prompt formulation and rubric design \citep{sclar2024quantifying, zhuo-etal-2024-prosa}, and can exhibit preferences for familiar styles or self-similar outputs \citep{NEURIPS2024_7f1f0218,wataoka2025selfpreferencebiasllmasajudge}. In our experiments, this brittleness appears clearly: while LLM judges perform well in assessing deductive mathematical reasoning, on open-ended clinical reasoning they over-accept generated segments.

We address this gap with a reference-free structural evaluation framework (audit) for LLM-generated reasoning traces. The framework looks at the step-by-step reasoning generated by an LLM as segments, labels local premise--target relations with NLI relations, and organizes them into a hypergraph. A deterministic backward AND--OR search algorithm then identifies which segments are \textsc{Supported}, \textsc{Unsupported}, or \textsc{Orphan}, turning evaluation into a granular audit of how the response grounds its own reasoning. \textbf{The framework has three components:}
 
\paragraph{Local NLI as the learning signal.}
We use NLI \citep{bowman2015largeannotatedcorpuslearning,havaldar2025entailedlinesincorporatingimplication} to label individual
premise--target relations with one of four labels: \textsc{Entailment}
for strong deductive support, \textsc{Implied} for weaker abductive
support, \textsc{Neutral} for insufficient information on relation determination, and
\textsc{Contradiction} for conflict. This four-way scheme extends
standard NLI with a separate label for abductive support, which is
the dominant inferential mode in open-ended reasoning, often conflated
with neutrality in three-way NLI.
 
\paragraph{NLI hypergraph as the global structure.}
\label{para:nli}
Reasoning targets often depend on the joint contribution of several
earlier segments, in a way no pairwise NLI judgement can capture. Therefore, we represent each trace as an NLI-labelled hypergraph, where
hypernodes denote candidate premise \emph{sets} and hyperedges denote
directed support from those sets to a target statement.
 
\paragraph{Backward AND--OR search as the audit decision.}
For each segment \(S_j\), we temporarily treat it as the \textbf{\emph{sink}} \textit{which refers to the current target segment being audited, not necessarily the final answer}. Starting from this target, we perform deterministic backward search over the hypergraph. The search alternates between an OR choice and an AND requirement: OR, because a segment may have several candidate supporting hypernodes; AND, because once a hypernode is chosen, all of its premises must themselves be grounded.

A segment is labelled \textsc{Supported} when a complete grounded support tree is recovered back to the context or context-grounded leaves. It is labelled \textsc{Unsupported} when local support edges exist but none can be expanded into a complete grounded tree. It is labelled \textsc{Orphan} when a segment has no incoming support edges or an ungrounded leaf.

% The latter two distinguish,
% \emph{floating} claims that are introduced without inferential anchoring
% from \emph{collapsing chains} which are locally plausible but rest on
% ungrounded foundations.

% The algorithm setup only requires the NLI label connection between the segments after which our deterministic white box approach showcases how the segment is connected in the tree and why its unfounded.

To evaluate the framework in a realistic open-ended setting, we introduce \textsc{UroReason}, a real-case medical reasoning benchmark for auditing LLM-generated clinical reasoning. This domain was chosen as it is of great clinical importance yet constitutes a very small part of LLM training and is not saturated unlike most reasoning datasets. The dataset contains 40 clinical cases consisting of corresponding evidence collected by practising physicians. For each case, an LLM generates a recommendation and reasoning trace for the next best clinical step, and physicians annotate the trace at the statement level. Unlike exam-style medical QA \citep{pal2022medmcqalargescalemultisubject,zuo2025medxpertqa}, \textsc{UroReason} provides evidence for the auditor rather than merely giving an answer: it can test whether an evaluator can detect errors, unsupported claims, hedging, and partial truths in LLM generated clinical reasoning. We also use Hard2Verify \citep{pandit2025hard2verifysteplevelverificationbenchmark} to compare evaluation behaviour across deductive mathematical reasoning. Hence,we make the following core contributions:
\begin{enumerate}[nosep]
 % Maria to continue from here
\item \textbf{A reference-free Evaluation Paradigm.} We introduce an evaluation framework for the quality assessment of LLM outputs. It converts an LLM-generated reasoning trace into an NLI-labelled hypergraph and audits each
statement through deterministic backward AND--OR search. The framework
requires no gold reference answers and is domain independent.
\item \textbf{\textsc{UroReason}:} A real life medical cases based benchmark with physician-labelled LLM reasoning traces, annotated at each reasoning step, for auditing open-ended clinical question answering.

 \item \textbf{Detailed experiments} evaluating the effectiveness of our reference-free reasoning based evaluation framework based on UroReason and Hard2Verify.

\item \textbf{A new failure mode for LLM-as-judge evaluation.}
We show that direct LLM judges can severely over-trust open-ended clinical reasoning, with specificity dropping to \(0.04\)--\(0.41\) on \textsc{UroReason}. Our NLI-hypergraph method mitigates this failure by decomposing evaluation into local inference labeling and deterministic grounded search.

\end{enumerate}
 
% \paragraph{Structural finding.}
% A side-effect of the audit is interpretive: it reveals that the shift
% from deductive maths to open-ended clinical reasoning is a
% \emph{structural} shift, not just a topical one. Mathematical traces
% are entailment-driven and produce deeper support trees; clinical
% traces are dominated by implied and multi-premise support, with
% shallower trees and substantially higher rates of floating claims.
% The audit discovers this distinction without being told which domain
% a trace comes from, and the difference in failure modes it exposes
% (\S\ref{sec:structural}, \S\ref{sec:catching}) is consistent with the
% deductive/abductive distinction across both panels.

\section{Related Work}

\paragraph{LLM-as-judge evaluation.}
LLM-as-judge evaluation has become a practical approach for assessing open-ended generations because it doesn't need exact-reference matching and can score free-form responses \citep{zheng2023judging}. However, judge-based evaluation is sensitive to prompt formulation, formatting, and rubric design \citep{sclar2024quantifying, zhuo-etal-2024-prosa}. Reliable use requires careful standardization, bias mitigation, and consistency checks \citep{NEURIPS2025_829e8f32}. For high-stakes domains like medicine, this level of uncertainty is risky and can reduce the adoption on a wider scale.

\noindent \textbf{Step-level verification and process supervision.}
Step-level verification addresses some limitations of final-answer evaluation by assessing intermediate reasoning steps, and has become central to process supervision and reasoning evaluation \citep{lightman2023letsverifystepstep,zheng-etal-2025-processbench}. However, much of the current literature is concentrated in mathematical or exam-style reasoning. Benchmarks such as MR-Ben \citep{NEURIPS2024_d81cb1f4}, ProcessBench \citep{zheng-etal-2025-processbench}, PRMBench \citep{song-etal-2025-prmbench} are largely derived from math-heavy sources. This setting encourages a primarily deductive view of verification, where each step is expected to follow from previous steps through explicit inference. This does not fully capture open-ended reasoning in complex domains. In medicine, law, and scientific analysis, a response often moves from incomplete evidence to plausible hypotheses, recommendations, or explanations. Such reasoning is frequently abductive rather than deductive.

% : a claim may be clinically or legally motivated without being entailed by a single preceding step. We therefore use mathematical reasoning as a deductive control setting, but focus on whether evaluation methods remain informative when reasoning becomes open-ended and abductive.

\noindent \textbf{Complex-domain reasoning benchmarks.}
Recent work has begun extending reasoning evaluation beyond mathematics. REVEAL studies relevance, attribution, and logical correctness in open-domain reasoning \citep{jacovi-etal-2024-chain}. Medical reasoning resources such as Med-PRM \citep{yun-etal-2025-med}, and MedThink-Bench \citep{zhou2025automatingexpertlevelmedicalreasoning} move toward step-level or rationale-based clinical evaluation. Other benchmarks target legal, biomedical, scientific, and narrative reasoning \citep{shojaee2025llmsrbenchnewbenchmarkscientific,chlapanis-etal-2025-greekbarbench}. %These efforts show that complex-domain reasoning evaluation is an important emerging direction. 
However, many existing resources remain tied to benchmark-style inputs and multiple-choice sources, which limits their ability to capture realistic open-ended QA.

% In such settings, complete references are expensive to construct and may not cover the variability of valid reasoning paths. Our work targets this gap by auditing the structure of generated reasoning traces directly.

\noindent \textbf{NLI as a reasoning primitive.}
NLI provides a practical primitive for decomposing reasoning into local premise--conclusion relations. Textual entailment was introduced through the Recognising Textual Entailment challenges \citep{pavlick-kwiatkowski-2019-inherent} and later scaled through SNLI and MNLI \citep{bowman2015largeannotatedcorpuslearning,williams-etal-2018-broad}. Abductive NLI extends this view by modelling inference toward plausible explanations rather than strict entailment alone \citep{bhagavatula2019abductive}. NLI has also been used for explanation and reasoning verification \citep{valentino-etal-2021-natural,quan-etal-2024-verification}. This makes NLI a natural formalism for our setting. Deduction can be approximated through entailment-like support, while abduction can be represented through weaker implied support.

\section{Methodology}
\label{sec:methodology}

\begin{figure*}[t]
    \centering
    % Replace 'figures/my_algorithm.pdf' with your actual file path
    \includegraphics[width=0.95\textwidth]{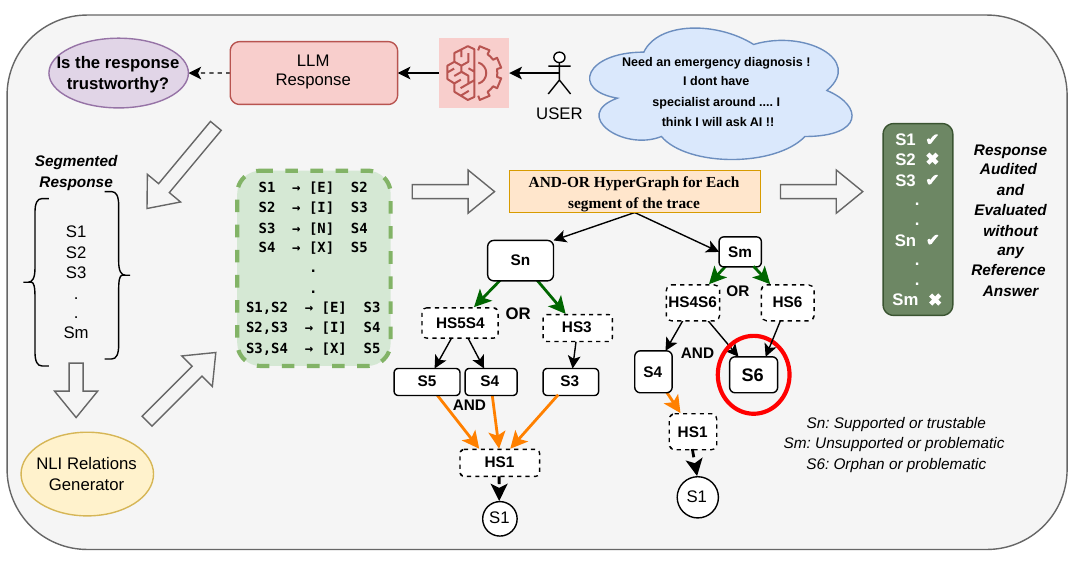}
    
    \caption{
\textbf{NLI-hypergraph audit framework.}
The response is segmented into reasoning units, candidate premise--target relations are labelled with NLI, and the resulting relations are organized into a hypergraph. A backward AND--OR search then composes local support relations to determine whether each segment is grounded in the generated trace, yielding audit labels such as \textsc{Supported}, \textsc{Unsupported}, and \textsc{Orphan}.
Orange arrows refer to the implied relation, and Green refers to the entailment relation. }
    \label{fig:main_algorithm}
\end{figure*}

We propose a reference-free method for auditing rationales generated by language models when providing a response to open-ended question answering (see Figure \ref{fig:main_algorithm}). Instead of checking every statement against external evidence, we ask whether each statement is supported by the model's own earlier reasoning. This lets us identify which parts of a rationale accompanying an output are grounded, which are weakly supported, and which appear without enough support. A statement is grounded if it can be connected to earlier parts of the response through clear local inference relations. It is non-grounded if it lacks support, conflicts with earlier content, or depends on another statement that could not be justified.

The framework has two stages. First, we convert the trace into an NLI-labelled hypergraph, where edges encode local premise--conclusion relations between earlier segments and later targets. Second, we apply deterministic graph analysis to this hypergraph, where we filter noisy support relations, recover grounded support trees through backward AND--OR search, and assign statement-level audit labels.
% This design keeps the learned component local: NLI is used only to judge premise--target relations, while the global audit is performed by an explicit structural procedure.

% This formulation distinguishes singleton support from compositional support: \(S_a\) and \(S_b\) may be insufficient individually, while \(\{S_a,S_b\}\) may jointly support \(S_j\). 

\subsection{Representation: Reasoning Trace as an NLI Hypergraph}

Assume a rationale/reasoning trace is generated by an LLM to justify a response to a question, in the form of a sequence of textual segments.
\[
T = (S_1, S_2, \ldots, S_n),
\]
where \(S_1\) denotes the context, question, or problem statement, and \(S_2,\ldots,S_n\) are generated reasoning segments. For each non-context segment \(S_j\), the framework predicts a binary structural label \(\hat{y}_j \in \{0,1\}\), where \(\hat{y}_j=1\) indicates that \(S_j\) is grounded in the recovered support structure and \(\hat{y}_j=0\) indicates that it is contradicted, unsupported, or disconnected.

A purely linear chain of pairwise relations is too restrictive for open-ended reasoning. Later segments often depend on several earlier observations jointly, rather than on a single preceding step. We therefore define support over candidate premise sets. For each target segment \(S_j\), we consider premise sets as
\[
P \subseteq \{S_1,\ldots,S_{j-1}\},
\qquad
m_{\min} \leq |P| \leq m_{\max},
\]
where \(P\) is a candidate set of earlier segments. An NLI model classifies the relation from \(P\) to \(S_j\), producing a labelled hyperedge
\[
e = (P, S_j, \ell).
\]

The resulting structure is a directed labelled hypergraph
\[
G=(V,E),
\]
where \(V\) contains the original segments and premise-set hypernodes, and \(E\) contains NLI-labelled relations from premise sets to target segments (See \S\ref{para:nli} for the set of NLI labels).

\subsection{Support Edge Selection}
\label{sec:edge_selection}

The raw NLI hypergraph contains all scored premise--target relations, many of which are weak, redundant, or unsuitable for grounding. We therefore prune the graph before running the AND--OR search algorithm.

\noindent \textbf{Pruning:} \textsc{Entailment} and \textsc{Implied} are treated as candidate support edges. \textsc{Neutral} is discarded as insufficient support. \textsc{Contradiction} is not used as support, but is retained separately as negative evidence for the target segment. We apply two deterministic pruning rules. First, \textsc{Implied} edges are kept only when the closest premise is within a fixed distance \(\tau\) of the target, reflecting the locality of abductive support. \textsc{Entailment} bypasses this filter. Second, when several premise sets support the same target, we remove redundant or dominated alternatives and prefer simpler or stronger candidates. The same \(\tau\) and rules are used across both domains; details and ablations are in Appendix~\ref{app:edge_selection}. The pruned graph is then passed to the AND--OR search algorithm.

\subsection{Grounded AND--OR Support Search Algorithm}

Support-edge selection yields local relations \(P \rightarrow S_j\), but local support alone is not enough. We need to make sure every premise in \(P\) must also be grounded. We therefore treat each segment \(S_j\) as a candidate sink and run backward to build a hypergraph representation.

We view the selected hypergraph as a bipartite AND--OR graph with two node types: segment nodes \(S_j\), and hypernodes \(H_P\), where \(H_P\) represents a premise set \(P=\{S_{a_1},\ldots,S_{a_k}\}\). A support edge \(P \rightarrow S_j\) is traversed backward as
\[
S_j \rightarrow H_P \rightarrow \{S_{a_1},\ldots,S_{a_k}\}.
\]
The first transition is an OR decision: a segment may have multiple candidate supporting hypernodes, and any one of them may explain the segment. The second transition is an AND decision: once \(H_P\) is selected, all premise segments in \(P\) must be recursively grounded.

% \begin{figure}[t]
% \centering
% \begin{tikzpicture}[
%     seg/.style={draw, rounded corners, minimum width=0.72cm, minimum height=0.45cm, align=center},
%     hyp/.style={draw, dashed, rounded corners, minimum width=0.70cm, minimum height=0.45cm, align=center},
%     every node/.style={font=\scriptsize},
%     >=stealth
% ]

% % Sink
% \node[seg] (s8) at (2.8,0) {$S_8$};

% % Hypernodes
% \node[hyp] (h23) at (1.35,-1.05) {$H_{\{2,3\}}$};
% \node[hyp] (h5)  at (4.25,-1.05) {$H_{\{5\}}$};

% % Premise segments
% \node[seg] (s2) at (0.55,-2.15) {$S_2$};
% \node[seg] (s3) at (2.15,-2.15) {$S_3$};
% \node[seg] (s5) at (4.25,-2.15) {$S_5$};

% % Edges
% \draw[->] (s8) -- (h23);
% \draw[->] (s8) -- (h5);

% \draw[->] (h23) -- (s2);
% \draw[->] (h23) -- (s3);
% \draw[->] (h5) -- (s5);

% % Labels
% \node at (2.8,-0.58) {\textbf{OR}};
% \node at (1.35,-1.65) {\textbf{AND}};
% \node at (4.25,-1.65) {\textbf{AND}};

% \end{tikzpicture}
% \caption{Backward AND--OR support search.}
% % A sink segment \(S_8\) may have alternative supporting hypernodes, giving an OR choice. Once a hypernode is selected, all premise segments inside it must be recursively grounded, giving an AND requirement.
% \label{fig:andor_search}
% \end{figure}

The search alternates between these two operations. At a segment node, the algorithm chooses among alternative supporting hypernodes. At a hypernode, it expands to all premise segments that constitute the hypernode. A branch terminates successfully when it reaches a grounded leaf.

In our setting, grounded leaves are the context segment \(S_1\). Fig. \ref{fig:audit_labels_tree} shows the first expansion process.
% A support tree is valid and grounded only if every branch reaches a grounded leaf.

\subsection{Segment-Level Audit Labels}

The grounded search assigns each non-context segment a structural audit label. A segment is labelled \textsc{Supported} if at least one complete grounded support tree is found leading from it to grounded leaf nodes (see Fig.~\ref{fig:audit_labels_tree} ) . It is labelled \textsc{Unsupported} if incoming support edges exist, but every candidate expansion fails to produce a complete grounded tree. It is labelled \textsc{Orphan} if no usable incoming support edge is available when the segment is treated as a sink. If contradiction evidence is detected, the segment is treated as a negative grounding case. We map \textsc{Supported} to \(\hat{y}=1\), and \textsc{Unsupported}, \textsc{Orphan}, and contradiction cases to \(\hat{y}=0\). These labels indicate whether a segment is structurally grounded in the generated trace; they do not claim external factual correctness. For efficiency, we use bounded deterministic search, prioritizing stronger NLI labels and closer premise--target relations. Fig.~\ref{fig:audit_labels_tree} illustrates the difference between a valid grounded tree and a failed candidate expansion.

\begin{figure}[t]
\centering
\begin{tikzpicture}[
    >=Stealth,
    x=1cm,
    y=1cm,
    seg/.style={
        draw,
        rounded corners,
        minimum width=0.72cm,
        minimum height=0.40cm,
        align=center,
        font=\scriptsize
    },
    hyp/.style={
        draw,
        dashed,
        rounded corners,
        minimum width=0.95cm,
        minimum height=0.42cm,
        align=center,
        font=\scriptsize
    },
    note/.style={
        font=\tiny,
        align=center,
        inner sep=1pt
    }
]

% ======================================================
% Panel A
% ======================================================
\node[font=\scriptsize\bfseries] at (1.6,0.55) {(a) Valid expansion};

\node[seg] (a_s8) at (1.6,0) {$S_8$};
\node[note] at (1.6,0.34) {\textsc{Supported}};

\node[hyp] (a_h56) at (1.6,-0.95) {$H_{\{5,6\}}$};

\node[seg] (a_s5) at (0.55,-1.85) {$S_5$};
\node[seg] (a_s6) at (2.65,-1.85) {$S_6$};

\node[note] at (0.55,-2.28) {context-\\grounded leaf};
\node[note] at (2.65,-2.28) {context-\\grounded leaf};

\draw[->] (a_s8) -- (a_h56);
\draw[->] (a_h56) -- (a_s5);
\draw[->] (a_h56) -- (a_s6);

\node[note] at (1.6,-0.48) {OR};
\node[note] at (1.6,-1.38) {AND};

% ======================================================
% Panel B
% ======================================================
\node[font=\scriptsize\bfseries] at (5.4,0.55) {(b) Failed expansion};

\node[seg] (b_s9) at (5.4,0) {$S_9$};
\node[note] at (5.4,0.34) {\textsc{Unsupported}};

\node[hyp] (b_h47) at (5.4,-0.95) {$H_{\{4,7\}}$};

\node[seg] (b_s4) at (4.35,-1.85) {$S_4$};
\node[seg] (b_s7) at (6.45,-1.85) {$S_7$};

\node[note] at (4.35,-2.28) {context-\\grounded leaf};
\node[note] at (6.45,-2.28) {\textsc{Orphan}};

\draw[->] (b_s9) -- (b_h47);
\draw[->] (b_h47) -- (b_s4);
\draw[->] (b_h47) -- (b_s7);

% failure mark on orphan branch
\draw[thick] (6.25,-1.65) -- (6.65,-2.05);
\draw[thick] (6.25,-2.05) -- (6.65,-1.65);

\node[note] at (5.4,-0.48) {OR};
\node[note] at (5.4,-1.38) {AND};

% ======================================================
% Bottom explanation
% ======================================================
\node[note] at (3.5,-3.05) {
Valid tree: all AND branches ground. \quad
Failed tree: one AND branch reaches an orphan.
};

\end{tikzpicture}
\caption{Segment-level grounding outcomes. In (a), sink \(S_8\) is \textsc{Supported} because every premise required by the selected hypernode reaches a context-grounded leaf. In (b), \(S_9\) has an incoming support candidate, but the AND condition fails because one required premise, \(S_7\), is an \textsc{Orphan}. Therefore no complete grounded tree is recovered and \(S_9\) is labelled \textsc{Unsupported}.}
\label{fig:audit_labels_tree}
\end{figure}

% \subsection{Reasoning-Forest Diagnostics}

% In addition to the final audit labels, we compute diagnostics over the recovered support trees. These diagnostics are not used to change the prediction; they are used to characterize the reasoning flow and identify failure modes.

% First, we count how often each segment appears in grounded support trees versus ungrounded candidate trees. This distinguishes segments that participate primarily in validated reasoning flows from segments that mainly appear in failed or floating support structures. Second, we deduplicate structurally redundant trees for profiling. Specifically, if one tree's hyperedge set is strictly contained in another tree's hyperedge set, the smaller tree is counted as redundant for the purpose of estimating the number of distinct reasoning flows.

% We also report diagnostics for negative labels. For orphan nodes, we distinguish cases where no incoming evidence exists from cases where evidence existed but was removed by locality filtering, local edge selection, or neutral NLI labels. This helps identify whether an error is due to the structural search or to upstream NLI relation quality.

% These diagnostics make the audit interpretable: a false positive supported segment can be traced to the support tree that accepted it, while a false negative orphan or unsupported segment can be traced to missing, filtered, or ungrounded support.

\subsection{Scope and Interpretation}

The framework audits structural trust within a generated response. It identifies how segments are organized, which claims are supported by earlier claims, and which parts of the trace are contradicted, unsupported, or disconnected. It does not perform external factual verification. If an early premise is false but later reasoning follows coherently from it, later segments may still be marked as structurally supported.

This scope is intentional. In open-ended question answering, exhaustive references may be unavailable or expensive to construct, and LLM-based judges can be opaque. Our method provides an intermediate audit layer: it converts a generated response into an inspectable reasoning structure and exposes where support is present, missing, contradictory, or unstable.

\section{Experimental Setup}
\label{sec:experimental_setup}

We evaluate our framework in two reasoning settings: an established deductive mathematical benchmark
as a controlled baseline, and a new open-ended clinical reasoning
dataset as an out-of-distribution test where evaluators face realistic abductive reasoning. Together these settings allow us to determine the generalisability of our reasoning-based reference-free evaluation framework.

\paragraph{Datasets}
For mathematical reasoning, we use Hard2Verify\citep{pandit2025hard2verifysteplevelverificationbenchmark}, an established step-level verification benchmark for Olympiad mathematical reasoning. We randomly sample 100 questions to form our deductive control setting.

\begin{figure}[t]
\centering
\includegraphics[width=\columnwidth]{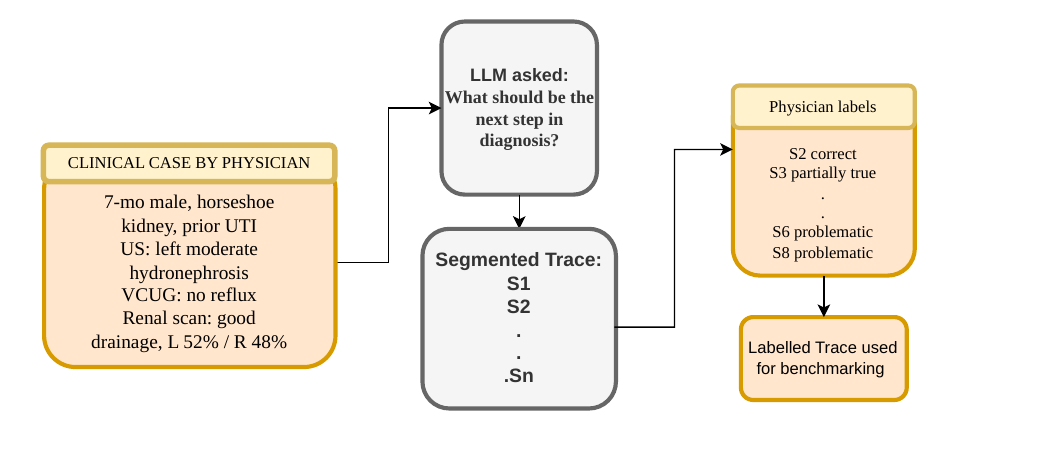}
\caption{
Overview of \textsc{UroReason}. Real clinical cases are used to elicit LLM-generated next-step recommendations and reasoning traces. These traces are segmented into reasoning units and annotated by physicians, producing a benchmark for evaluating step-level reasoning-audit methods.
}
\label{fig:uroreason_dataset}
\end{figure}

For open-ended real-world reasoning, we introduce \textsc{UroReason}, a medical reasoning-audit benchmark developed with practising clinicians. It is built from physician-curated real patient scenarios, each containing multiple sources of evidence including clinical history, examination findings, investigations, and clinical impressions. Given a case, an LLM is prompted to recommend the next best clinical step and generate a reasoning trace to justify the response. Human experts then label the trace at the segment level, allowing us to evaluate whether step-level judges or audit frameworks can identify unsupported, incomplete, or clinically problematic reasoning in open-ended question answering pertaining to real-world medical reasoning.

The final dataset contains 40 clinical cases and 305 labelled reasoning segments. To our knowledge, \textsc{UroReason} is the first step-level verification benchmark focused on open-ended real-world medical reasoning, moving beyond multiple-choice or exam-style medical QA.

\paragraph{Generating reasoning traces.}
We generate traces using Qwen3-4B-Think-2507 \citep{qwen3technicalreport} where the LLM is prompted to recommend the next best clinical step and justify its reasoning. 

% Generation prompts are provided in Appendix~X.

\paragraph{Trace segmentation and annotation.}
Each generated rationale is split into reasoning units (segments) using natural punctuation boundaries in the model output. Two physicians, one senior and one junior, then annotate each segment in the context of the full clinical case. Segments are labeled as correct, incorrect, hedging/inconclusive, or partially true.

For evaluation, we collapse these labels into a binary audit setting: segments that are clearly correct and clinically appropriate are mapped to \(y=1\), while incorrect, hedging/inconclusive, and partially true segments are mapped to \(y=0\). This conservative mapping reflects the goal of \textsc{UroReason}: to identify segments that should be trusted without review. In clinical reasoning, a partially true or inconclusive statement can still mislead by omitting or overstating a finding, or supporting an inappropriate recommendation; such segments should therefore be flagged rather than endorsed.

\paragraph{Evaluation paradigms.}
We compare two evaluation paradigms. In the \emph{LLM-as-judge} paradigm, an LLM directly classifies each reasoning segment as trustworthy or problematic. This represents the standard direct-verification baseline. In our \emph{structural audit} paradigm, the LLM is not used as the final verifier. Instead, it is used only to provide local NLI labels for directed premise--target relations \((P \rightarrow S_j)\). These local labels are then composed by the deterministic AND--OR grounding algorithm described in \S\ref{sec:methodology}.

We evaluate the same four open-weight model families in both paradigms: GPT-OSS-120B, GPT-OSS-20B, Gemma~4-31b-it, and Qwen3-30B. We also include 4 frontier API models: Claude Sonnet 4.6, DeepSeek V4 flash, and Gemini-3.1-Flash-Lite and GPT 5.4-mini as direct judges, to test whether the observed patterns hold beyond open-weight models. The NLI prompt is domain-general and held fixed across both mathematics and medicine (we do not perform per-domain prompt tuning). Thus, variation across NLI sources is treated as a robustness test for our framework. All judge prompts, NLI prompts, and decoding parameters are provided in Appendix~X.

\paragraph{Metrics.}
We report sensitivity (TPR, recall on $y=1$), specificity (TNR,
recall on $y=0$), Class~0 F1 ($F1_0$), and balanced F1. Following Hard2Verify, balanced
F1 is the harmonic mean of TPR and TNR:
\[
\text{Bal F1}
   = \frac{2 \cdot \text{TPR} \cdot \text{TNR}}
          {\text{TPR} + \text{TNR}},
\]
giving a single summary statistic that is robust to class imbalance
and that penalises evaluators with degenerate one-sided behaviour
(uniform acceptance or uniform flagging).

\section{Results and Analysis}

\label{sec:results}

\definecolor{vanillaColor}{HTML}{993C1D}
\definecolor{algoColor}{HTML}{185FA5}
\definecolor{posDelta}{HTML}{0F6E56}
\definecolor{negDelta}{HTML}{A32D2D}
\definecolor{rowAlt}{HTML}{F7F6F2}
 
\newcommand{\vh}[1]{\textcolor{vanillaColor}{#1}}
\newcommand{\ah}[1]{\textcolor{algoColor}{#1}}
\newcommand{\dpos}[1]{\textcolor{posDelta}{\textbf{#1}}}
\newcommand{\dneg}[1]{\textcolor{negDelta}{#1}}

We organize the results around three questions. First, how do direct LLM judges behave when evaluation shifts from deductive mathematical reasoning to open-ended clinical reasoning? Second, what changes when the same LLMs are used only as local NLI sources inside our hypergraph audit? Third, what does the recovered graph reveal about how reasoning fails across domains?

Section~\ref{sec:main_results} reports the main performance comparison. Section~\ref{sec:structural} analyzes audit-label distributions, showing that mathematical failures often resemble broken chains, while clinical failures more often involve unanchored claims. Section~\ref{sec:case_studies} grounds this pattern in a qualitative case study.
 
% ===============================================================
\subsection{Main Results}
\label{sec:main_results}
% ===============================================================
\begin{table*}[t]
    \centering
    
    % ================= LEFT SIDE: SUBTABLE 1a =================
    \begin{subtable}[c]{0.53\textwidth}
        \centering
        \renewcommand{\arraystretch}{1.1} 
        
        % This guarantees the table never spills into the margin
        \resizebox{\linewidth}{!}{%
        \begin{tabular}{l ccc !{\color{black!25}\vrule} ccc !{\color{black!25}\vrule} c}
        \toprule
        & \multicolumn{3}{c}{\textcolor{vanillaColor}{\textbf{LLM as Step Level Evaluator}}}
        & \multicolumn{3}{c}{\textcolor{algoColor}{\textbf{Ours (NLI)}}}
        & \\
        \cmidrule(lr){2-4} \cmidrule(lr){5-7}
        \textbf{LLM}
         & TPR & TNR & Step-level Bal
         & TPR & TNR & Step-level Bal
         & $\Delta$ Bal \\
        \midrule
        \multicolumn{8}{l}{\textit{Maths --- Hard2Verify, 100 traces, 916 segments}} \\
        \midrule
        GPT-OSS-120B & \vh{0.964} & \vh{0.530} & \vh{0.684} & \ah{0.529} & \ah{0.788} & \ah{0.633} & \dneg{$-0.051$} \\
        \rowcolor{rowAlt} GPT-OSS-20B & \vh{0.913} & \vh{0.535} & \vh{0.674} & \ah{0.569} & \ah{0.763} & \ah{0.652} & \dneg{$-0.022$} \\
        Gemma 4 & \vh{0.974} & \vh{0.365} & \vh{0.531} & \ah{0.633} & \ah{0.725} & \ah{0.676} & \dpos{$+0.145$} \\
        \rowcolor{rowAlt} Qwen3-30B & \vh{0.970} & \vh{0.306} & \vh{0.465} & \ah{0.484} & \ah{0.790} & \ah{0.600} & \dpos{$+0.135$} \\
        
        \arrayrulecolor{black!25}\midrule\arrayrulecolor{black}
        
        \multicolumn{8}{l}{\textit{Medical --- UroReason, 40 cases, 305 segments}} \\
        \midrule
        GPT-OSS-120B & \vh{0.950} & \vh{0.207} & \vh{0.340} & \ah{0.525} & \ah{0.641} & \ah{0.577} & \dpos{$+0.237$} \\
        \rowcolor{rowAlt} GPT-OSS-20B & \vh{0.975} & \vh{0.103} & \vh{0.187} & \ah{0.563} & \ah{0.586} & \ah{0.574} & \dpos{$+0.387$} \\
        Gemma 4 & \vh{0.988} & \vh{0.228} & \vh{0.370} & \ah{0.519} & \ah{0.607} & \ah{0.559} & \dpos{$+0.189$} \\
        \rowcolor{rowAlt} Qwen3-30B & \vh{0.956} & \vh{0.041} & \vh{0.079} & \ah{0.594} & \ah{0.510} & \ah{0.549} & \dpos{$+0.470$} \\
        \bottomrule
        \end{tabular}%
        }
        \caption{Paired comparison of evaluation paradigms.}
        \label{tab:sub_paradigm_compare}
    \end{subtable}%
    \hfill
    % ================= RIGHT SIDE: SUBTABLE 1b =================
    \begin{subtable}[c]{0.44\textwidth}
        \centering
        \renewcommand{\arraystretch}{1.1} 
        
        % This guarantees the table never spills into the margin
        \resizebox{\linewidth}{!}{%
        \begin{tabular}{l ccc !{\color{black!25}\vrule} ccc}
        \toprule
        & \multicolumn{3}{c}{\textbf{Hard2Verify}}
        & \multicolumn{3}{c}{\textbf{UroReason}} \\
        \cmidrule(lr){2-4} \cmidrule(lr){5-7}
        \textbf{Evaluator}
         & \textbf{TPR} & \textbf{TNR} & \textbf{Bal}
         & \textbf{TPR} & \textbf{TNR} & \textbf{Bal} \\
        \midrule
        
        Ours-best & 0.633 & 0.725 & 0.676 & 0.525 & \textbf{0.641} & \textbf{0.577} \\
        \rowcolor{rowAlt} GPT-5.4-mini & 0.563 & \textbf{0.857} & 0.679 & 0.819 & 0.434 & 0.568 \\
        Claude Sonnet 4.6 & 0.972 & 0.456 & 0.621 & 0.900 & 0.407 & 0.560 \\
        \rowcolor{rowAlt} Ours-worst & 0.484 & 0.790 & 0.600 & 0.519 & 0.510 & 0.549 \\
        \rowcolor{rowAlt} DeepSeek V4-Flash & 0.748 & 0.700 & \textbf{0.724} & 0.931 & 0.283 & 0.434 \\
        Gemini3.1-Flash & \textbf{0.989} & 0.222 & 0.362 & \textbf{0.975} & 0.152 & 0.263 \\
        
        \bottomrule
        \end{tabular}%
        }
        \caption{Cross-domain generalization gap.}
        \label{tab:sub_domain_comp}
    \end{subtable}
    
    % ================= MASTER CAPTION =================
    \vspace{4pt}
    \caption{\textbf{Comprehensive evaluation results.} \textbf{(a)} Shows the impact of shifting from a vanilla LLM-as-judge to our NLI-based structural audit using identical open-weights. Positive $\Delta$ values indicate an increase in Balanced F1. \textbf{(b)} Highlights the generalization gap across domains for closed-source models.}
    \label{tab:master_results}
\end{table*}
\noindent \textbf{Direct judges fail under open-ended clinical reasoning.}
\label{sec:sota_collapse}
% ---------------------------------------------------------------------
Table~\ref{tab:master_results}(b) compares strong direct LLM judges across Hard2Verify and \textsc{UroReason}. On Hard2Verify, several judges remain competitive, with the best model reaching \(0.724\) balanced-F1. On \textsc{UroReason}, direct judging becomes less reliable: TPR remains high, but TNR drops sharply, reaching only \(0.434\) for GPT-5.4-mini, \(0.407\) for Claude Sonnet4.6, and \(0.152\) for Gemini3.1-Flash. This high-TPR/low-TNR pattern indicates over-acceptance: judges often mark problematic, hedged, or partially true clinical segments as trustworthy.

This motivates \textsc{UroReason} as a stress test for evaluator reliability. Direct judging can appear effective on deductive verification while failing in open-ended medical reasoning, where plausibility and correctness are harder to separate \textit{(The variance in the results of DeepseekV4 is a prime example of this)}. Our NLI-hypergraph audit is more stable in this setting, with the best configuration achieving the highest \textsc{UroReason} balanced-F1 and the weakest configuration remaining competitive with frontier judges.

\noindent \textbf{The proposed NLI-hypergraph audit improves reliability.}
\label{sec:rescue_effect}
% ---------------------------------------------------------------------

\noindent Table~\ref{tab:master_results} (a) compares direct judging with our NLI-hypergraph method using the same four open-weight LLMs: GPT-OSS-120B, GPT-OSS-20B, Gemma~4, and Qwen3-30B. The open-weight judges reproduce the same false-trust pattern on \textsc{UroReason}: TPR remains high (\(0.956\)--\(0.988\)), but TNR collapses to \(0.041\)--\(0.228\), giving balanced-F1 scores of \(0.079\)--\(0.370\). When the same models are used only for local NLI labelling and their outputs are processed by our graph algorithm, balanced-F1 rises to \(0.549\)--\(0.577\). Qwen3-30B improves from \(0.079\) to \(0.549\), GPT-OSS-20B from \(0.187\) to \(0.574\), Gemma~4 from \(0.370\) to \(0.559\), and GPT-OSS-120B from \(0.340\) to \(0.577\).

The gains are not limited to \textsc{UroReason}. On Hard2Verify, the method substantially improves weaker direct judges: Gemma~4 rises from \(0.531\) to \(0.676\), and Qwen3-30B from \(0.465\) to \(0.600\). The strongest direct judges remain competitive on math, as expected for explicit deductive chains and existing open benchmarks. Overall, the framework converts local NLI signals into a more robust evaluator, with the largest gains appearing when step into out of domain tasks.
% \paragraph{The structural composition is what stabilises evaluation.}
% The variance contrast in Table~\ref{tab:main_results} is the central
% empirical claim of this paper. Across the four open-weight LLMs,
% vanilla Balanced~F1 spans 0.219 points on maths and 0.291 points on
% UroReason. As NLI sources for our algorithm, the same four LLMs span
% only 0.076 and 0.028 points respectively --- a roughly three-fold
% compression on maths and ten-fold on UroReason. Different LLMs
% disagree on many individual NLI judgements, but those disagreements
% average out under the deterministic composition. What stabilises
% evaluation across LLMs and domains is not the choice of LLM, it is
% the structural composition step.

\begin{table}[t]
\centering
\renewcommand{\arraystretch}{1.15}
% This line mathematically guarantees it will not spill into the next column
\resizebox{\columnwidth}{!}{%
\begin{tabular}{l ccc !{\color{black!25}\vrule} ccc}
\toprule
& \multicolumn{3}{c}{\textbf{Maths Domain (\%)}}
& \multicolumn{3}{c}{\textbf{Medical Domain (\%)}} \\
\cmidrule(lr){2-4} \cmidrule(lr){5-7}
\textbf{LLM Evaluator}
 & \textbf{\textsc{Orph.}} & \textbf{\textsc{Unsup.}} & \textit{Ratio}
 & \textbf{\textsc{Orph.}} & \textbf{\textsc{Unsup.}} & \textit{Ratio} \\
\midrule

GPT-OSS-120B  
 & 32.0 & 26.4 & \textit{1.2$\times$} 
 & 46.2 & 6.6  & \textit{7.0$\times$} \\

\rowcolor{rowAlt} GPT-OSS-20B   
 & 29.3 & 22.9 & \textit{1.3$\times$} 
 & 40.3 & 6.9  & \textit{5.8$\times$} \\

Gemma 4       
 & 31.4 & 28.7 & \textit{1.1$\times$} 
 & 42.6 & 9.8  & \textit{4.3$\times$} \\

\rowcolor{rowAlt} Qwen3-30B     
 & 26.0 & 33.2 & \textit{0.8$\times$} 
 & 37.7 & 6.2  & \textit{6.1$\times$} \\

\bottomrule
\end{tabular}%
} % End of resizebox
\caption{\textbf{Global Audit Failure Modes.} Comparison of \textsc{Orphan} and \textsc{Unsupported} segment classifications produced by our framwork. The ratio (\textsc{Orphan} / \textsc{Unsupported}) highlights the structural contrast: Medical reasoning fails primarily by floating without anchors, while Mathematical reasoning fails through collapsed logic chains.}
\label{tab:audit_failures}
\end{table}

% ===============================================================
\subsection{Structural Signatures Across Domains}
\label{sec:structural}
% ===============================================================

The previous section shows that the NLI-hypergraph method is more stable than direct judging, especially on \textsc{UroReason}. We next examine what the method reveals about the structure of the generated reasoning traces. We focus on the audit outcomes produced by the graph search: \textsc{Grounded}, \textsc{Orphan}, and \textsc{Unsupported}. These labels are structural diagnostics, not factual truth labels. A segment marked \textsc{Orphan} or \textsc{Unsupported} is not necessarily false but it only means that it's not grounded in the generated trace.

\noindent \textbf{Medical reasoning is more orphan-dominant.}
Table~\ref{tab:audit_failures} shows a consistent difference between Hard2Verify and \textsc{UroReason}. In math, \textsc{Orphan} and \textsc{Unsupported} occur at comparable rates across NLI sources: \textsc{Orphan} ranges from \(26.0\%\) to \(32.0\%\), while \textsc{Unsupported} ranges from \(22.9\%\) to \(33.2\%\). This suggests that mathematical failures often occur inside attempted derivational chains: some local support exists, but recursive grounding fails.

In \textsc{UroReason}, the pattern changes. \textsc{Orphan} dominates across all NLI sources, ranging from \(37.7\%\) to \(46.2\%\), while \textsc{Unsupported} remains much lower, between \(6.2\%\) and \(9.8\%\). This reflects the structure of open-ended medical reasoning. LLMs often introduce multiple hypotheses, plausible observations, or recommendation-like claims without explicitly anchoring each one in prior case evidence. These claims may sound clinically plausible, but if the trace does not provide a recoverable support path, the graph search marks them as \textsc{Orphan}.

\noindent \textbf{The pattern is stable across NLI sources.}
The same qualitative contrast appears for all four NLI sources. Although the exact percentages vary, every configuration shows a more orphan-dominant profile on \textsc{UroReason} than on Hard2Verify. This stability matters because it suggests that the observed difference is not an artefact of one local labeller. The graph-level outcomes consistently separate the two domains: mathematical reasoning more often fails through broken chains, while clinical reasoning more often fails through unanchored claims.

\subsection{Qualitative Case Study: Local Plausibility vs. Grounding}
\label{sec:case_studies}

Table~\ref{tab:medical_case_trees} in the appendix illustrates why direct LLM judges over-accept open-ended clinical reasoning. The generated trace is not irrelevant or incoherent. It uses real case facts: the patient has a mild varicocele, no obvious emergency features, and meatal stenosis. However, the reasoning moves from these local observations to a management recommendation that the physician marks as clinically misdirected. In particular, the response moves toward reassurance and no further imaging, while the physician notes that the next step should consider voiding stream assessment, voiding diary, possible urethromeatoplasty, and follow-up imaging. A full tree representation created by our algorithm is showcased at Figure \ref{fig:appendix_diagram}

The audit exposes where this reasoning flow breaks. \(S_4\), which identifies the varicocele as clinically relevant, is marked \textsc{Supported}; the trace contains enough local evidence for this observation. By contrast, \(S_6\) and \(S_7\) are marked \textsc{Orphan}: they introduce plausible claims about meatal stenosis and self-resolving varicocele-related discomfort, but no recoverable support path is found for them in the trace. The recommendation-level claims in \(S_8\) and \(S_{10}\) are marked \textsc{Unsupported}: candidate support exists, but it depends on weakly grounded or orphaned premises and does not compose into a grounded path to the appropriate next step.

This example showcases the types of errors that our framework can capture. Medical reasoning can sound plausible while still being poorly grounded. A direct judge may reward fluent medical prose and locally reasonable statements, but miss that the global support flow is broken. The NLI-hypergraph audit enforces that the response grounds its own reasoning and captures: where support is present, where it disappears, and where final recommendations rely on ungrounded premises.

\section{Conclusion}
We present a reference-free framework for auditing LLM-generated rationales for output responses. Instead of relying on final-answer matching or direct LLM-as-judge scores, our method analyses response segments by modelling local NLI relations, and using grounded hypergraph search to identify which segments are \textsc{Supported}, \textsc{Unsupported}, or \textsc{Orphan}.

We introduce \textsc{UroReason}, a physician-annotated benchmark for open-ended medical reasoning. Our results show that direct LLM judges often over-accept fluent medical reasoning, while our NLI-hypergraph audit provides a more balanced signal across both deductive (Hard2Verify) and open-ended real-world reasoning (\textsc{UroReason}). The structural analysis further shows that math and medical reasoning fail differently: in the former errors often appear as broken support chains, whereas traces of the latter often contain plausible but unanchored claims.

Future work will extend this framework by training smaller NLI relation models, to avoid dependency on LLM and expanding stepwise auditing to additional medical and high-stakes domains. Overall, our findings suggest that trustworthy evaluation of open-ended QA should inspect how reasoning steps connect, not only whether the final answer appears plausible.

% \paragraph{NLI labels provide supporting evidence.}
% The raw NLI label distributions are consistent with this interpretation, but we treat them as supporting evidence rather than ground truth. Across NLI sources, Hard2Verify contains more \textsc{Entailment}-like support, while \textsc{UroReason} contains substantially more \textsc{Implied} support. This matches the expected distinction between deductive mathematical reasoning and abductive clinical reasoning, where conclusions are often plausible rather than strictly entailed. Full NLI label distributions and tree-depth statistics are provided in Appendix~\ref{app:structural_full}.

\section*{Limitations}
\label{sec:limitations}

Our work has several limitations. First, the framework currently relies on LLMs to produce local NLI labels. This is a practical choice: existing NLI models are not yet reliable enough for the kinds of long, domain-specific, and multi-premise reasoning relations studied here, especially in mathematical and clinical traces. However, this means that the audit is not fully independent of LLM behaviour. We mitigate this by restricting the LLM to local premise--target relation labelling and using deterministic hypergraph search for the final decision, but future work should train smaller specialized NLI models for reasoning-trace auditing.

Second, we do not study test-time scaling. All judge and NLI-label runs use fixed prompting and greedy decoding. Exploring self-consistency, repeated sampling, debate-style judging, or verifier ensembles may improve both direct LLM-as-judge baselines and local NLI labelling. We leave this analysis to future work due to compute constraints for open-weight models and cost constraints for frontier API models.

Third, \textsc{UroReason} is intentionally specialized and relatively small. It contains 40 real paediatric urology cases and 305 labelled reasoning segments. The dataset size reflects the cost of collecting realistic clinical cases and obtaining high-quality physician annotations. The annotations were produced with practising clinicians, including a senior paediatric urologist with more than 25 years of experience and a junior physician in the same clinical setting. This gives the benchmark high clinical relevance, but it also limits domain coverage. Future releases should expand to more specialties, institutions, and case types.

% Fourth, our audit evaluates structural grounding within the generated trace, not external factual correctness. A segment may be structurally grounded but still clinically wrong if the trace begins from an incorrect assumption, and a clinically correct statement may be flagged if the model does not explicitly support it. This is by design: the framework is intended as a reference-free review signal, not a replacement for expert judgement or retrieval-based factual verification. A natural extension is to combine trace grounding with external medical knowledge sources or retrieval systems.

Finally, our focus is open-ended reasoning rather than multiple-choice QA. Many existing medical benchmarks condition model reasoning on predefined answer options, which can constrain the hypothesis space and reduce the kinds of free-form hallucination, hedging, and partial-truth behaviour seen in real user interactions. \textsc{UroReason} targets this more realistic setting, but future work should test whether the same structural audit principles transfer to other open-ended high-stakes domains such as law, finance, and policy reasoning.

% \section*{Acknowledgments}

% This document has been adapted
% by Steven Bethard, Ryan Cotterell and Rui Yan
% from the instructions for earlier ACL and NAACL proceedings, including those for
% ACL 2019 by Douwe Kiela and Ivan Vuli\'{c},
% NAACL 2019 by Stephanie Lukin and Alla Roskovskaya,
% ACL 2018 by Shay Cohen, Kevin Gimpel, and Wei Lu,
% NAACL 2018 by Margaret Mitchell and Stephanie Lukin,
% Bib\TeX{} suggestions for (NA)ACL 2017/2018 from Jason Eisner,
% ACL 2017 by Dan Gildea and Min-Yen Kan,
% NAACL 2017 by Margaret Mitchell,
% ACL 2012 by Maggie Li and Michael White,
% ACL 2010 by Jing-Shin Chang and Philipp Koehn,
% ACL 2008 by Johanna D. Moore, Simone Teufel, James Allan, and Sadaoki Furui,
% ACL 2005 by Hwee Tou Ng and Kemal Oflazer,
% ACL 2002 by Eugene Charniak and Dekang Lin,
% and earlier ACL and EACL formats written by several people, including
% John Chen, Henry S. Thompson and Donald Walker.
% Additional elements were taken from the formatting instructions of the \emph{International Joint Conference on Artificial Intelligence} and the \emph{Conference on Computer Vision and Pattern Recognition}.

\bibliography{custom}

% ===============================================================
\appendix
% ===============================================================

\section{Implementation Details}
\label{app:implementation}

This appendix provides implementation details for the support-edge selection rules and the prompts used in our experiments. These details are included for reproducibility; the main paper describes the conceptual framework and reports the primary results.

\subsection{Support Edge Selection}
\label{app:edge_selection}

Before running grounded support-tree search, we prune the raw NLI hypergraph. The raw graph contains all scored premise--target relations, including weak, redundant, or noisy edges. The goal of support-edge selection is to keep only relations that can plausibly contribute to grounding a target segment.

We order NLI labels by support strength:
\[
\textsc{Contra} < \textsc{Neutral} < \textsc{Implied} < \textsc{Entail}.
\]
\textsc{Entailment} and \textsc{Implied} are treated as candidate support edges. \textsc{Neutral} edges are not used for support-tree construction. \textsc{Contradiction} edges are retained separately as negative evidence for the target segment.

For each target \(S_j\), we compare multi-premise support edges against their singleton components. If a multi-premise edge provides stronger support than its components, the combination is retained. If a simpler edge already provides strong support, the redundant combination is removed. If combining premises weakens otherwise strong support, the target is flagged as unstable. Repeated weak support is handled conservatively to avoid introducing noisy abductive paths.

We also apply a locality prior to weak support. \textsc{Entailment} edges bypass this prior, while \textsc{Implied} edges may be removed when their closest premise, farthest premise, or premise span exceeds a fixed threshold. These thresholds are treated as implementation parameters and are ablated separately.

\begin{table}[h]
\centering
\small
\setlength{\tabcolsep}{3pt}
\renewcommand{\arraystretch}{1.1}
\begin{tabular}{@{}p{0.24\columnwidth}p{0.30\columnwidth}p{0.38\columnwidth}@{}}
\toprule
\textbf{Pattern} & \textbf{Typical form} & \textbf{Action} \\
\midrule

\textsc{Conjunctive}
& Combination improves over components
& Keep the multi-premise edge. \\

\textsc{Redundant}
& Simpler strong support already exists
& Keep the simpler edge; remove the redundant combination. \\

\textsc{Asymmetric}
& One component carries stronger support
& Retain the usable support edge. \\

\textsc{Degraded}
& Strong components weaken under combination
& Flag the target as unstable; remove the weakened combination. \\

\textsc{Consistently Weak}
& Weak evidence remains weak
& Remove the weak cluster conservatively. \\

\textsc{Contradicted}
& Contradiction evidence exists
& Mark the target as contradicted. \\

\bottomrule
\end{tabular}

\caption{
Implementation-level support-edge selection patterns. These patterns are used to clean the NLI hypergraph before support-tree search and are not treated as final semantic audit labels.
}
\label{tab:edge_selection_appendix}
\end{table}

% ===============================================================
\section{Prompt Templates}
\label{app:prompts}
% ===============================================================

We provide the exact prompt templates used for direct LLM-as-judge baselines and for NLI relation labelling. Variables populated dynamically during inference are shown in monospaced font, e.g., \texttt{\{problem\}}.

\subsection{Direct LLM-as-Judge Prompts}
\label{app:judge_prompts}

The following prompts are used for the vanilla LLM-as-judge baselines. In this setting, the LLM directly predicts whether each reasoning step is correct or incorrect. The prompts differ slightly across math and clinical domains to match the terminology of each task, but both require a step-level verdict.

\begin{promptbox}{Baseline Evaluator Prompt: Maths Domain}
The following is a math problem and a solution (split into steps, enclosed with tags and indexed from 0):

\textbf{[Math Problem]}
\texttt{\{problem\}}

\textbf{[Solution]}
\texttt{\{steps\}}

Your task is to review and critique the solution step-by-step.
For each step, determine if it is correct or incorrect. A correct step is one where all of the content is correct, and is logically consistent with all previous steps and information given in the problem.

An incorrect step is one where the content is incorrect, or is not logically consistent with all previous steps and information given in the problem, or is based on an error in a previous step.

\textbf{Important:} Any step that contains or is based on an error is considered incorrect. That is, if the error is carried forward from a previous step or is based on an error in the previous step, consider the step incorrect. Provide reasoning for your correctness determinations.

There are exactly \texttt{\{expected\_n\}} solution steps, indexed from Step 0 to Step \texttt{\{last\_idx\}}.

Do not give a verdict for the math problem itself.
Give exactly one verdict for each solution step.
Do not skip any step.
Do not add extra steps.

Please use the following format to return your answer:

\textbf{Reasoning:}
Step 0: \textless your reasoning\textgreater
Step 1: \textless your reasoning\textgreater
...

\textbf{Verdict:}
Step 0: \textless yes or no\textgreater
Step 1: \textless yes or no\textgreater
...
Step \texttt{\{last\_idx\}}: \textless yes or no\textgreater

\textbf{Rules:}
\begin{itemize}
    \setlength{\itemsep}{0pt}
    \item Use \texttt{yes} if the step is correct.
    \item Use \texttt{no} if the step is incorrect.
    \item Each verdict line must start with \texttt{Step \textless index\textgreater:}
    \item Each verdict must be exactly \texttt{yes} or \texttt{no}.
    \item Do not give a verdict for the math problem itself.
    \item Do not add any text after the final verdict line.
\end{itemize}
\end{promptbox}

\vspace{1em} % Adds some space between the two boxes

% ================= MEDICAL BASELINE =================
\begin{promptbox}{Baseline Evaluator Prompt: Medical Domain}
The following is a clinical case and a diagnostic / management reasoning trace split into steps, enclosed with tags and indexed from 0:

\textbf{[Clinical Case]}
\texttt{\{problem\}}

\textbf{[Reasoning Trace]}
\texttt{\{steps\}}

Your task is to evaluate each reasoning step.

There are exactly \texttt{\{n\_steps\}} reasoning steps.

For each step, label it as CORRECT or INCORRECT.

A step should be labeled \texttt{CORRECT} if it is clinically appropriate, supported by the case, and logically consistent.\\
A step should be labeled \texttt{INCORRECT} if it is clinically inappropriate, unsupported, contradicted by the case, or logically inconsistent.

Provide brief reasoning for your labels.

Your final verdict must be a comma-separated list of exactly \texttt{\{n\_steps\}} labels. Each label must be either \texttt{CORRECT} or \texttt{INCORRECT}. The labels must correspond to the reasoning steps in order from step 0 to step \texttt{\{last\_step\_idx\}}.

Use exactly this format:

\textbf{Reasoning:} \textless brief explanation\textgreater\\
\textbf{Verdict:} \textless comma-separated list of CORRECT or INCORRECT labels\textgreater

Do not include more or fewer than \texttt{\{n\_steps\}} labels.
\end{promptbox}

\subsection{NLI Relation Prompt}
\label{app:nli_prompt}

The following prompt is used in the first stage of our framework to label local premise--target relations. The prompt enforces directed local inference: the model must evaluate whether the premise \(P\) supports the proposed conclusion \(C\), without using global context unless it is explicitly included in \(P\).

\begin{promptbox}{NLI Relation Generator Prompt (One-to-One)}
Classify one directed relation in a reasoning trace.

Let P = PREMISE.\\
Let C = PROPOSED CONCLUSION.

Evaluate only the independent edge: \textbf{P $\rightarrow$ C}

Treat P as the only available premise.
\begin{itemize}
    \setlength{\itemsep}{0pt}
    \item Do not use any other reasoning step unless it is explicitly included inside P.
    \item Do not assume C is correct because it appears later in the solution.
    \item Do not judge whether C is a reasonable continuation of the full proof.
    \item Do not judge whether the full answer is correct.
    \item Do not judge whether C supports P.
    \item Do not rely only on shared words, symbols, topic, or chronological order.
\end{itemize}

\textbf{Ask:} ``Using only P, is C established, weakly motivated, unsupported, or incompatible?''

If C requires intermediate steps that are not present in P, do not label \texttt{entailment}.\\
If C depends on earlier assignments, cases, definitions, or conclusions not contained in P, do not label \texttt{entailment}.

\textbf{Choose exactly one label:}
\begin{description}
    \item[\texttt{entailment}] P locally establishes C.\\
    Use this only when C follows from P by a direct and valid local inference, such as a restatement, definition, calculation, application of a stated rule, or immediate consequence.

    \item[\texttt{implied}] P gives weak but meaningful support for C.\\
    Use this when C is a plausible, natural, or motivated next step from P, but P does not establish C as true. Use this when P suggests C, but C still needs an intermediate bridge not present in P.

    \item[\texttt{neutral}] P does not give enough local support for C.\\
    Use this when C adds information, assumptions, restrictions, cases, claims, assignments, configurations, or conclusions that are not established by P. Use this when C requires intermediate reasoning not present in P. Use this when C is compatible with P but not justified by P. Use this when C turns one possible branch or case into a broader conclusion without justification.

    \item[\texttt{contradiction}] C is incompatible with P.\\
    Use this only when P and C cannot both be true in the same reasoning state. Do not use contradiction merely because C is unsupported.
\end{description}

\textbf{Important checks:}
\begin{itemize}
    \setlength{\itemsep}{0pt}
    \item P alone establishes C $\rightarrow$ \texttt{entailment}
    \item P alone motivates C but does not establish it $\rightarrow$ \texttt{implied}
    \item P alone does not support C enough $\rightarrow$ \texttt{neutral}
    \item P conflicts with C $\rightarrow$ \texttt{contradiction}
\end{itemize}

\textbf{PREMISE P:}\\
\texttt{\{premise\}}

\textbf{PROPOSED CONCLUSION C} (\texttt{\{conclusion\_id\}})\textbf{:}\\
\texttt{\{conclusion\}}

Return EXACTLY this format:

\textless rationale\textgreater\\
Briefly explain the local relation from P alone to C and reflect on the label you give.\\
\textless /rationale\textgreater

\textless answer\textgreater\\
\texttt{\{conclusion\_id\}}: \texttt{entailment} / \texttt{implied} / \texttt{neutral} / \texttt{contradiction}\\
\textless /answer\textgreater
\end{promptbox}

% ===============================================================
\section{Additional Structural Diagnostics}
\label{app:structural_full}
% ===============================================================

\begin{figure}[h]
\centering
\includegraphics[width=0.95\columnwidth]{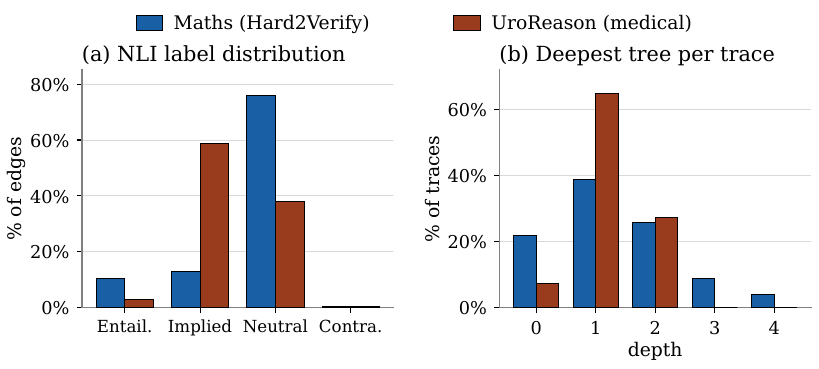}
\caption{
Structural signatures of mathematical and medical reasoning traces. The recovered graph shows a shift from more entailment-driven support in mathematical reasoning to more implied and abductive support in open-ended clinical reasoning.
}
\label{fig:structural_appendix}
\end{figure}

Figure~\ref{fig:structural_appendix} provides additional structural diagnostics comparing mathematical and clinical reasoning traces. These diagnostics supplement the audit-label analysis in Section~\ref{sec:structural}.

% ===============================================================
\section{Qualitative Case Studies}
\label{app:qualitative_cases}
% ===============================================================

This section provides additional qualitative examples supporting the analysis in Section~\ref{sec:case_studies}. The main paper reports a compact case analysis; here we include fuller traces and annotations.

\subsection{Full Clinical Case Study}
\label{app:case_study_full}

Table~\ref{tab:medical_case_full} provides the full \textsc{UroReason} case discussed in Section~\ref{sec:case_studies}. The case involves follow-up after testicular pain, mild left varicocele, and meatal stenosis. We report the generated reasoning segment, physician annotation, binary gold label, and audit label assigned by our framework.

\begin{table*}[t]
\centering
\scriptsize
\setlength{\tabcolsep}{3pt}
\renewcommand{\arraystretch}{1.15}
\begin{tabular}{p{0.045\textwidth}p{0.42\textwidth}p{0.28\textwidth}p{0.07\textwidth}p{0.12\textwidth}}
\toprule
\textbf{Step} 
& \textbf{Generated reasoning segment} 
& \textbf{Physician annotation} 
& \textbf{Gold} 
& \textbf{Audit label} \\
\midrule

\(S_2\)
& The patient presents for follow-up after a recent episode of testicular pain that began in early October and lasted approximately one and a half weeks, with gradual resolution over time, consistent with a self-limited, non-acute condition.
& Correct.
& 1
& \textsc{Supported} \\

\(S_3\)
& The constant but progressively improving pain, together with a similar episode two years ago that resolved without intervention, suggests a benign vascular or mechanical cause rather than torsion or infection.
& Uncertain: explanation is incomplete and underspecified; unclear what is meant by a ``benign vascular or mechanical cause.''
& 0
& \textsc{Supported} \\

\(S_4\)
& The mild grade 1--2 left varicocele, confirmed by ultrasound, is notable because varicoceles can occasionally cause dull aching pain, especially with standing or prolonged activity.
& Correct.
& 1
& \textsc{Supported} \\

\(S_5\)
& The pain was mild and transient, and the absence of fever, dysuria, hematuria, or sudden severe pain makes a significant urological emergency unlikely.
& Correct.
& 1
& \textsc{Orphan} \\

\(S_6\)
& Meatal stenosis is noted, but the response states that it is not directly related to testicular pain, while also noting possible urinary-flow effects.
& Uncertain: incomplete; narrowed meatus can create backpressure with retrograde flow into the epididymis.
& 0
& \textsc{Orphan} \\

\(S_7\)
& Given symptom resolution, benign varicocele, and absence of concerning features, the response concludes that the likely diagnosis is mild, self-resolving varicocele-related discomfort.
& Uncertain: incomplete; varicocele pain is typically intermittent and recurrent, not self-limiting.
& 0
& \textsc{Orphan} \\

\(S_8\)
& The response recommends reassurance and periodic monitoring, stating that the varicocele is likely benign and does not require immediate intervention.
& Incorrect: should observe voiding stream and consider urethromeatoplasty or voiding diary.
& 0
& \textsc{Unsupported} \\

\(S_9\)
& The response recommends documenting meatal stenosis and considering further evaluation only if urinary hesitancy, incomplete emptying, or difficulty initiating urination develops.
& Correct but incomplete: decision to proceed with urethromeatoplasty is driven by quality of urinary stream.
& 1
& \textsc{Orphan} \\

\(S_{10}\)
& The response states that no further imaging or surgical referral is indicated at this time.
& Incorrect: annual ultrasound is needed to track testis volume.
& 0
& \textsc{Unsupported} \\

\bottomrule
\end{tabular}

\caption{
Full qualitative \textsc{UroReason} case. The example illustrates why clinical reasoning is difficult to audit: several problematic segments are plausible, incomplete, or partially true rather than obviously false. \textsc{Orphan} indicates that no recoverable incoming support is found in the generated trace; \textsc{Unsupported} indicates that candidate support exists but does not compose into a grounded reasoning path. These labels are structural review signals, not external clinical truth judgements.
}
\label{tab:medical_case_full}
\end{table*}

\subsection{Recovered Support Trees for the Clinical Case}
\label{app:case_tree_breakdown}

Table~\ref{tab:medical_case_trees} summarizes the support structures recovered for the clinical case in Table~\ref{tab:medical_case_full}. We report only the interpretable support paths used by the audit, rather than the full raw hypergraph.

\subsection{Recovered Support Trees for the Clinical Case}
\label{app:case_tree_breakdown}

Table~\ref{tab:medical_case_trees} summarizes the support structures recovered for the clinical case in Table~\ref{tab:medical_case_full}. The goal is to show where the audit finds grounded reasoning and where the reasoning flow breaks. We report the interpretable support paths used by the audit rather than the full raw hypergraph.

\begin{table*}[t]
\centering
\small
\setlength{\tabcolsep}{4pt}
\renewcommand{\arraystretch}{1.15}
\begin{tabular}{p{0.08\textwidth}p{0.18\textwidth}p{0.25\textwidth}p{0.41\textwidth}}
\toprule
\textbf{Target} 
& \textbf{Audit outcome} 
& \textbf{Recovered / candidate support} 
& \textbf{Interpretation} \\
\midrule

\(S_2,S_3\)
& \textsc{Supported}
& \(S_1 \rightarrow S_2\), \(S_1 \rightarrow S_3\)
& The early pain-history interpretation is connected to the case context. This establishes the initial local grounding of the trace, although \(S_3\) is physician-labelled problematic because its explanation is underspecified. \\

\(S_4\)
& \textsc{Supported}
& \(S_1 \rightarrow S_4\); \(\{S_2,S_3\}\rightarrow S_4\)
& The audit recovers a valid support point: the mild varicocele is grounded in the case context and in the early pain-history discussion. This shows that the framework does not reject the trace wholesale; it identifies the part that is locally supported. \\

\(S_5\)
& \textsc{Orphan}
& No retained grounded support tree
& This is the first breakpoint. The claim that a urological emergency is unlikely may be clinically acceptable, but the audit does not recover a support path that connects it cleanly to the earlier grounded trace. It is therefore flagged as a conservative review point. \\

\(S_6\)
& \textsc{Orphan}
& No retained grounded support tree
& The trace introduces meatal stenosis but treats it as not directly related to pain. The audit identifies this as an unanchored claim: it enters the reasoning without a recoverable support path. This aligns with the physician concern that the mechanism is incomplete. \\

\(S_7\)
& \textsc{Orphan}
& No retained grounded support tree
& The response shifts from a grounded varicocele observation to a diagnosis-like claim about self-resolving varicocele discomfort. The audit marks this as a second breakpoint: the claim sounds plausible, but it is not grounded by the preceding trace. \\

\(S_8\)
& \textsc{Unsupported}
& Candidate: \(\{S_4,S_5\}\rightarrow S_8\)
& The audit finds why the recommendation is tempting: \(S_4\) makes the varicocele locally relevant and \(S_5\) suggests no emergency. However, this support path does not ground the correct management step. Since \(S_5\) is not itself grounded and the path does not address the physician-relevant next step, the recommendation remains unsupported. \\

\(S_9\)
& \textsc{Orphan}
& No retained grounded support tree
& The statement about documenting meatal stenosis is physician-labelled correct but incomplete. The audit flags it because the trace does not recoverably connect it to the earlier reasoning. This illustrates the conservative nature of the audit. \\

\(S_{10}\)
& \textsc{Unsupported}
& Candidates: \(\{S_5,S_6\}\rightarrow S_{10}\), \(\{S_7,S_8\}\rightarrow S_{10}\), \(\{S_8,S_9\}\rightarrow S_{10}\)
& The final no-imaging/no-referral claim has local candidate supports, but each path depends on earlier orphaned or unsupported segments. The audit therefore identifies the final collapse: the recommendation is not merely unsupported in isolation, but built from a weak reasoning path. \\

\bottomrule
\end{tabular}

\caption{
Recovered and failed support structures for the qualitative clinical case. The audit identifies the main breakpoints in the reasoning flow: early varicocele-related observations are locally grounded, but later diagnostic and management claims become \textsc{Orphan} or \textsc{Unsupported}. This shows how a fluent clinical trace can move from plausible local facts to an unsupported final recommendation.
}
\label{tab:medical_case_trees}
\end{table*}

\section{Experimental Setup}
\label{app:experimental_setup}

\subsection{LLM Inference Setup}
\label{app:llm_inference_setup}

All LLM-as-judge baselines and NLI relation-labelling prompts are decoded greedily with temperature \(0.0\). For open-weight models, we serve the models using vLLM on a single H100 96GB GPU. The maximum model length is set to \(8192\), and the maximum number of output tokens is set to \(4096\).

For API-based models accessed through OpenRouter, we use a maximum context length of \(4096\) for both Hard2Verify and \textsc{UroReason}, again with temperature \(0.0\). Direct LLM-as-judge baselines produce segment-level labels end-to-end. In contrast, for our framework, the LLM is used only to generate local NLI relations; the final segment-level audit labels are produced by deterministic hypergraph search.

\subsection{Hypergraph Audit Configuration}
\label{app:hypergraph_config}

Table~\ref{tab:hypergraph_config} reports the main configuration for the proposed NLI-hypergraph audit. The implementation uses deterministic candidate ordering and bounded search to ensure reproducibility and avoid combinatorial expansion.

\begin{table}[h]
\centering
\small
\setlength{\tabcolsep}{4pt}
\renewcommand{\arraystretch}{1.15}
\begin{tabular}{p{0.42\columnwidth}p{0.20\columnwidth}p{0.30\columnwidth}}
\toprule
\textbf{Component} & \textbf{Setting} & \textbf{Role} \\
\midrule

Premise size
& \(1\)--\(2\)
& Allows singleton and two-premise support. \\

Context premises
& Enabled
& Allows \(S_1\) to ground early reasoning. \\

Support labels
& \textsc{Entail}, \textsc{Implied}
& Candidate edges for tree search. \\

Negative labels
& \textsc{Neutral}, \textsc{Contradiction}
& Neutral is excluded; contradiction is retained as negative evidence. \\

Premise prefilter
& Gap-based
& Removes distant weak \textsc{Implied} support edges. \\

Closest-gap threshold
& \(3\)
& Locality threshold for weak support. \\

Tree search
& Best-first
& Deterministic bounded support-tree construction. \\

Max tree depth
& \(100\)
& Upper bound on recursive search. \\

Max trees per sink
& \(50\)
& Prevents tree explosion. \\

Max subtrees per node
& \(50\)
& Bounds recursive branching. \\

\bottomrule
\end{tabular}

\caption{
Main configuration for the proposed NLI-hypergraph audit. These settings control support-edge construction, locality filtering, and bounded grounded tree search.
}
\label{tab:hypergraph_config}
\end{table}
\subsection{Hypergraph Search Configuration}
\label{app:hypergraph_config}

Table~\ref{tab:hypergraph_config} summarizes the main configuration used for our NLI-hypergraph audit. The configuration controls three stages: premise construction, locality-based edge filtering, and bounded support-tree search.

\subsection{Gap-Based Filtering and Best-First Search}
\label{app:gap_best_first}

\paragraph{Gap-based locality filtering.}
For a candidate support edge \(P \rightarrow S_j\), let \(P=\{S_{i_1},\ldots,S_{i_k}\}\) be the premise set and \(S_j\) the target segment, where all premises occur before the target: \(i_m < j\). We define three distance quantities:
\[
d_{\min}(P,S_j) = j - \max_{S_i \in P} i,
\]
\[
d_{\max}(P,S_j) = j - \min_{S_i \in P} i,
\]
\[
\operatorname{span}(P) = \max_{S_i \in P} i - \min_{S_i \in P} i.
\]
Here, \(d_{\min}\) is the distance from the target to its closest premise, \(d_{\max}\) is the distance to its farthest premise, and \(\operatorname{span}(P)\) measures how spread out the premises are.

We use these quantities as a locality prior for weak support. In the main configuration, \textsc{Implied} edges are removed when the closest premise is beyond the configured gap threshold:
\[
d_{\min}(P,S_j) \geq \tau.
\]
\textsc{Entailment} edges bypass this filter because they represent stronger support. The motivation is practical: abductive or weak support is more reliable when it is local, while distant \textsc{Implied} edges are more likely to reflect topical similarity rather than usable support. In our main setting, the gap filter is enabled with \(\tau=3\), while farthest-gap and premise-span thresholds are set high enough not to bind. The implementation exposes these as ablation parameters. 

\paragraph{Best-first grounded search.}
After filtering, each segment \(S_j\) is treated as a candidate sink. The goal is to determine whether \(S_j\) can be grounded by recursively tracing support edges backward to the context or to early context-grounded steps.

For each sink \(S_j\), the algorithm considers candidate incoming edges
\[
e = (P \rightarrow S_j),
\]
where \(P\) is a premise set. Candidate edges are sorted deterministically by the priority key
\[
\pi(e) =
\left(
-r(e),\;
d_{\min}(e),\;
|P|,\;
P
\right),
\]
where \(r(e)\) is the NLI label rank:
\[
r(\textsc{Entailment}) = 3,\qquad
r(\textsc{Implied}) = 2.
\]
Thus, the search first tries stronger NLI edges, then closer support edges, then smaller premise sets, and finally a stable index order over premises. This makes the search deterministic and reproducible.

For an edge \(P \rightarrow S_j\) to form a valid support tree, every premise \(S_i \in P\) must itself be grounded. The algorithm therefore recursively searches each premise. If all branches reach the context \(S_1\) or an accepted context-grounded early step, the tree is valid and \(S_j\) is labelled \textsc{Supported}. If no incoming support edge exists, \(S_j\) is labelled \textsc{Orphan}. If incoming candidate edges exist but every recursive expansion fails, \(S_j\) is labelled \textsc{Unsupported}. Search is bounded by maximum depth, maximum trees per sink, and maximum subtrees per node to prevent combinatorial blow-up.

\paragraph{Interpretation.}
These settings do not change the semantic meaning of the labels. They are practical constraints for turning a dense NLI hypergraph into a usable grounding graph. A segment is labelled \textsc{Supported} only if at least one complete grounded tree is found. It is labelled \textsc{Orphan} if no usable incoming support survives edge selection, and \textsc{Unsupported} if candidate support exists but cannot be recursively grounded.

\paragraph{Dataset statistics.}
\textsc{UroReason} contains 40 clinical cases and 305 labelled reasoning segments, with an average of 7.62 segments per case. Segment lengths are moderate, with a mean of 39.60 words and a median of 37 words per segment. At the raw annotation level, 160 segments are labelled correct (52.46\%), 74 incorrect (24.26\%), and 71 uncertain (23.28\%). For binary audit evaluation, we map only clearly correct segments to \(y=1\), while incorrect and uncertain segments are mapped to \(y=0\). This yields a balanced split of 160 trustworthy segments (52.46\%) and 145 problematic segments (47.54\%).

\begin{figure*}[t]
\centering
\includegraphics[
    width=0.95\textwidth,
    keepaspectratio
]{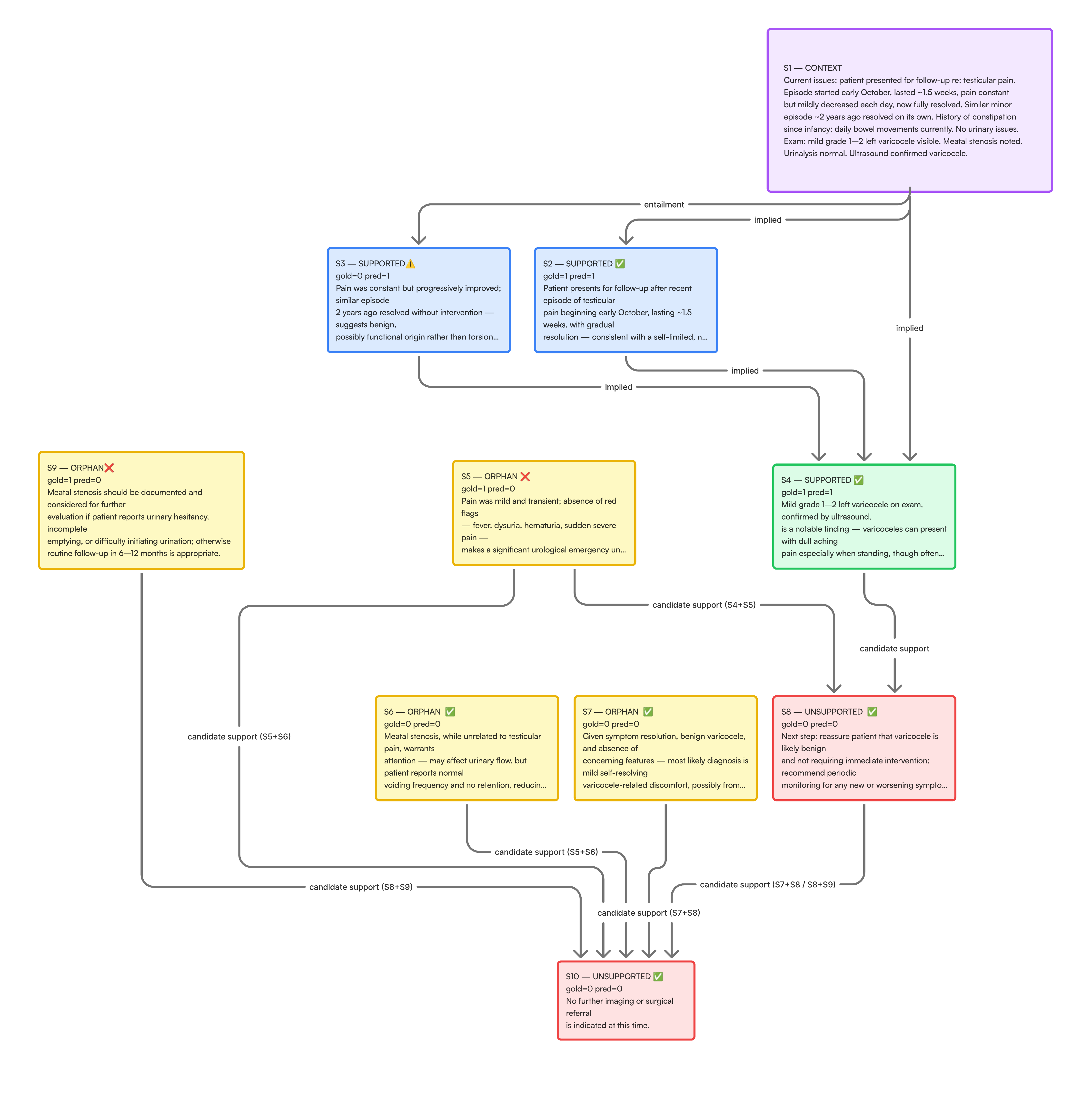}
\caption{
A break down representation of how the reasoning hypergraph looks like
}
\label{fig:appendix_diagram}
\end{figure*}

\begin{table}[t]
\centering
\small
\setlength{\tabcolsep}{5pt}
\renewcommand{\arraystretch}{1.1}
\begin{tabular}{l r}
\toprule
\textbf{Statistic} & \textbf{Value} \\
\midrule
Clinical cases & 40 \\
Labelled reasoning segments & 305 \\
Mean segments per case & 7.62 \\
Median segments per case & 7 \\
Min / max segments per case & 5 / 12 \\
Mean words per segment & 39.60 \\
Median words per segment & 37 \\
\bottomrule
\end{tabular}
\caption{
Summary statistics for \textsc{UroReason}. Segment counts exclude the clinical case context and include only generated reasoning units.
}
\label{tab:uroreason_stats}
\end{table}
\begin{table}[t]
\centering
\small
\setlength{\tabcolsep}{5pt}
\renewcommand{\arraystretch}{1.1}
\begin{tabular}{l r r}
\toprule
\textbf{Label} & \textbf{Count} & \textbf{Percent} \\
\midrule
Correct & 160 & 52.46 \\
Incorrect & 74 & 24.26 \\
Uncertain & 71 & 23.28 \\
\midrule
Trustworthy (\(y=1\)) & 160 & 52.46 \\
Problematic (\(y=0\)) & 145 & 47.54 \\
\bottomrule
\end{tabular}
\caption{
Annotation distribution in \textsc{UroReason}. For binary audit evaluation, only clearly correct segments are treated as trustworthy; incorrect and uncertain segments are treated as problematic.
}
\label{tab:uroreason_label_dist}
\end{table}

\section{Physician Annotation Guidelines}
\label{app:annotation_guidelines}

This section describes the guidelines provided to physician annotators for labelling the clinical reasoning traces in \textsc{UroReason}. Annotators were asked to evaluate each generated reasoning segment as if they were reviewing an explanation written by a fellow clinician. The goal was not to grade the final answer alone, but to identify which parts of the reasoning trace were clinically sound, incomplete, misleading, or incorrect.

\subsection{Annotation Task}

For each clinical case, annotators were shown the full patient scenario and an LLM-generated reasoning trace. The trace was divided into reasoning segments. Annotators were asked to judge each segment in the context of the full case and the preceding reasoning.

Annotators were instructed to answer the following question for each segment:

\begin{quote}
Is this reasoning segment clinically appropriate, supported by the case, and safe to treat as trustworthy in the context of the full explanation?
\end{quote}

They were also asked to provide a short comment whenever a segment was not clearly correct, explaining the clinical issue.

\subsection{Annotation Labels}

Each segment was assigned one of three labels: \textsc{Correct}, \textsc{Incorrect}, or \textsc{Uncertain}. The labels were defined as follows.

\paragraph{\textsc{Correct}.}
A segment should be labelled \textsc{Correct} if it is clinically appropriate, supported by the case information, and does not omit an important qualification that would change the interpretation. A correct segment may be a factual observation, a clinically reasonable inference, or an appropriate recommendation.

\paragraph{\textsc{Incorrect}.}
A segment should be labelled \textsc{Incorrect} if it contains a clinically wrong statement, contradicts the case, misinterprets a finding, recommends an inappropriate next step, or could mislead clinical decision-making.

\paragraph{\textsc{Uncertain}.}
A segment should be labelled \textsc{Uncertain} if it is not clearly false but is incomplete, vague, overconfident, hedged, insufficiently justified, or only partially true. Annotators were asked to use this label when a statement sounded plausible but lacked enough specificity or omitted an important clinical qualifier.

\subsection{General Annotation Principles}

Annotators were asked to follow these principles:

\begin{itemize}
    \item Judge each segment in the context of the full clinical case, not in isolation.
    \item Evaluate whether the segment is clinically justified by the provided history, examination, investigations, and impressions.
    \item Do not reward a segment only because it sounds fluent or plausible.
    \item Mark a segment as problematic if it omits a clinically important qualification.
    \item Mark a segment as problematic if it supports an inappropriate diagnosis, recommendation, or management plan.
    \item Use \textsc{Uncertain} for partially true or under-specified reasoning rather than forcing a binary correct/incorrect judgement.
    \item Provide a brief rationale for every \textsc{Incorrect} or \textsc{Uncertain} label.
\end{itemize}

\subsection{Examples of Label Decisions}

\paragraph{Correct segment.}
A statement that accurately identifies a finding present in the case and gives a clinically reasonable interpretation should be labelled \textsc{Correct}. For example, if the case reports a varicocele and the reasoning states that varicoceles can sometimes be associated with dull aching pain, this can be labelled correct if the statement is appropriately qualified.

\paragraph{Incorrect segment.}
A statement should be labelled \textsc{Incorrect} if it leads to an inappropriate management recommendation. For example, if the reasoning states that no further imaging is needed when follow-up imaging is clinically required, the segment should be labelled incorrect.

\paragraph{Uncertain segment.}
A statement should be labelled \textsc{Uncertain} if it is plausible but incomplete. For example, a segment that refers to a ``benign vascular or mechanical cause'' without specifying the mechanism or clinical relevance may be labelled uncertain because it lacks sufficient explanatory detail.

\subsection{Binary Mapping for Evaluation}

For evaluation, we collapse the physician labels into binary audit labels. Segments labelled \textsc{Correct} are mapped to \(y=1\). Segments labelled \textsc{Incorrect} or \textsc{Uncertain} are mapped to \(y=0\).

This conservative mapping reflects the intended use of \textsc{UroReason} as an audit benchmark. In clinical reasoning, a partially true, vague, or inconclusive statement can still mislead a downstream reader by omitting a necessary qualifier, overstating a finding, or supporting an inappropriate recommendation. Such segments should therefore be flagged for review rather than treated as trustworthy.
\end{document}